\documentclass{article}
\usepackage{ijcai16}

\usepackage{times}
\usepackage{helvet}
\usepackage{courier}

\usepackage{algorithm2e}
\usepackage{amsmath}
\usepackage{amsfonts}

\usepackage{comment}
\usepackage[it]{caption}
\usepackage{subcaption}
\usepackage{algorithmic,float}
\usepackage{graphicx}
\usepackage{booktabs}

\usepackage{wrapfig}

\newcommand{\citep}[1]{\cite{#1}}
\newcommand{\citenp}[1]{\citeauthor{#1} \citeyear{#1}}
\newcommand{\citet}[1]{\citeauthor{#1} (\citeyear{#1})}

\newcommand{\citeynp}[1]{\citeyear{#1}}
\newcommand{\maxscore}[1]{{\bf #1}}

\title{Deep Learning for Reward Design \\ to Improve Monte Carlo Tree Search in ATARI Games}
\author{Xiaoxiao Guo, Satinder Singh, Richard Lewis, Honglak Lee \\
University of Michigan, Ann Arbor \\
\{guoxiao,baveja,rickl,honglak\}@umich.edu}

\begin{document}

\maketitle

\begin{abstract}
Monte Carlo Tree Search (MCTS) methods have proven powerful in planning for sequential decision-making problems such as Go and video games, but their performance can be poor when the planning depth and sampling trajectories are limited or when the rewards are sparse. We present an adaptation of PGRD (policy-gradient for reward-design) for learning a reward-bonus function to improve UCT (a MCTS algorithm).  Unlike previous applications of PGRD in which the space of reward-bonus functions was limited to linear functions of hand-coded state-action-features, we use PGRD with a multi-layer convolutional neural network to automatically learn features from raw perception as well as to adapt the non-linear reward-bonus function parameters.   We also adopt a variance-reducing gradient method to improve PGRD's performance. The new method improves UCT's performance on multiple ATARI games compared to UCT without the reward bonus.  Combining PGRD and Deep Learning in this way should make adapting rewards for MCTS algorithms far more widely and practically applicable than before. 
\end{abstract}

\section{Introduction}

This paper offers a novel means of combining Deep Learning (DL; see~\citenp{bengio2009learning};~\citenp{schmidhuber2015deep} for surveys) and Reinforcement Learning (RL), with an application to ATARI games. There has been a flurry of recent work on combining DL and RL on ATARI games, including the seminal work using DL as a function approximator for Q-learning~\cite{mnih2015human}, the use of UCT-based planning to provide policy-training data for a DL function approximator~\cite{guo2014deep}, the use of DL to learn transition-models of ATARI games to improve exploration in Q-learning~\cite{oh2015action,stadie2015incentivizing}, and the use of DL as a parametric representation of policies to improve via policy-gradient approaches~\cite{schulman2015trust}.
%% This paper, on the other hand, 
%,SDMIA15-Hausknecht

In contrast, the work presented here uses DL as a function approximator to learn {reward-bonus functions} from experience to mitigate computational limitations in UCT~\cite{kocsis2006bandit}, a Monte Carlo Tree Search (MCTS) algorithm. In large-scale sequential decision-making problems, UCT often suffers because of limits on the number of trajectories and planning depth required to keep the method computationally tractable. %~\cite{15aaai-mnnuct}. 
%, improving its performance during repeated planning in ATARI games.
%We describe how this is accomplished below.   
The key contribution of this paper is a new method for improving the performance of UCT planning in such challenging settings, exploiting the powerful feature-learning capabilities of DL.

Our work builds on PGRD (policy-gradient for reward-design;~\citenp{sorg2010reward}), a method for learning  reward-bonus functions for use in planning. Previous applications of PGRD have been limited in a few ways: 1) they have mostly been applied to small RL problems; 2) they have required careful hand-construction of features that define a reward-bonus function space to be searched over by PGRD; 3) they have limited the reward-bonus function space to be a linear function of hand-constructed features; and 4) they have high-variance in the gradient estimates (this issue was not apparent in earlier work because of the small size of the domains). In this paper, we address all of these limitations by developing PGRD-DL, a PGRD variant that automatically learns features over raw perception inputs via a multi-layer convolutional neural network (CNN), and uses a variance-reducing form of gradient. We show that PGRD-DL can improve performance of UCT on multiple ATARI games. 

\section{Background and Related Work}
{\bf ALE as a challenging testbed for RL.}
The Arcade Learning Environment (ALE) includes an ATARI 2600 emulator and about 50 games \cite{bellemare2013arcade}. These games present a challenging combination of perception and  policy selection problems. All of the  games have the same high-dimensional observation screen, a 2D array of 7-bit pixels, 160 pixels wide by 210 pixels high.  The action space depends on the game but consists of at most 18 discrete actions.  Building agents to learn to play ATARI games is challenging due to the high-dimensionality and partial observability of the perceptions, as well as the sparse and often highly-delayed nature of the rewards. There are a number of approaches to building ATARI game players, including some that do not use DL
 (e.g.,~\citenp{bellemare2012sketch},~\citeynp{bellemare2012investigating},~\citeynp{bellemare2013bayesian},~\citenp{lipovetzky2015classical}). Among these are players that use UCT to plan with the ALE emulator as a model \cite{bellemare2013arcade}, an approach that we build on here.

{\bf \noindent Combining DL and RL.} 
Deep Learning is a powerful set of techniques for learning feature representations from raw input data in a compositional hierarchy, where low-level features are learned to encode low-level statistical dependencies (e.g. edges in images), and higher-level features encode higher-order dependencies of the lower-level features (e.g. object parts;~\citenp{lee2009convolutional}). Combining DL and RL methods together thus offers an opportunity to address the high-dimensional perception, partial observability and sparse and delayed reward challenges in building ATARI game playing agents. Indeed, as summarized above, recent work has started to do just that, though research into combining neural networks and RL is older than the recent interest in ALE (e.g.,~\citenp{schmidhuber1990online};~\citenp{tesauro1994td}).

\vspace*{0.5em}
{\bf \noindent UCT.}
UCT is widely used for planning in large scale sequential decision games.
%because the complexity of the action selection step is independent of the size of the state space (allowing it to be applied to ATARI games, for example).
It takes three parameters: the number of trajectories, the maximum-depth, and an exploration parameter. In general, the larger the trajectory and depth parameters, the better the performance, though at the expense of increased computation at each action selection. UCT computes a score for each possible action $a$ in state-depth pair $(s,d)$ as the sum of two terms: an exploitation term that is the Monte Carlo average of the discounted sum of rewards obtained from experiences with state-depth pair $(s,d)$ in the previous $k-1$ trajectories, and an exploration term that encourages sampling infrequently taken actions.\footnote{Specifically, the exploration term is $c\sqrt{\log{n(s,d)}/n(s,a,d)}$ where $n(s,d)$ and $n(s,a,d)$ are the number of visits to state-depth pair $(s,d)$, and of action $a$ in state-depth pair $(s,d)$ respectively in the previous $k-1$ trajectories, and $c$ is the exploration parameter.} 
UCT selects the action to simulate in order to extend the trajectory greedily with respect to the summed score. Once the number of sampled trajectories reaches  the parameter value, UCT returns the exploitation term for each action at the root node as its estimate of the utility of taking that action in the current state of the game. 

Although they have been successful in many applications, UCT and more generally MCTS methods also have limitations: their performance suffers when the number of trajectories and planning depth are limited relative to the size of the domain, or when the rewards are sparse, which creates a kind of needle-in-a-haystack problem for search. 

\vspace*{0.5em}
{\bf \noindent Mitigating UCT limitations.} Agent designers often build in heuristics for overcoming these UCT drawbacks. As one example, in the case of limited planning depth, the classical Leaf-Evaluation Heuristic (LEH)
%~\cite{shannonxxii} 
adds a heuristic value to the estimated return at the leaf states. However, the value to compensate for the missing subtree below the leaf is often not known in practice.  Many other approaches improve UCT by mechanisms for generalization of returns and visit statistics across states: e.g., the Transposition Tables of \citet{childs2008transpositions}, the Rapid Action Value Estimation of \citet{gelly2011monte}, the local homomorphisms of \citet{jiang2014improving}, the state abstractions of \citet{hostetler2014state}, and the local manifolds of \citet{15aaai-mnnuct}.  

In this paper, we explore learning a reward-bonus function to mitigate the computational bounds on UCT. Through deep-learned features of state, our method also provides a generalization mechanism that allows UCT to exploit useful knowledge gained in prior planning episodes. 

\vspace*{0.5em}
{\bf \noindent Reward design, optimal rewards, and PGRD.} ~\citet{singh-IMRL} proposed a framework of optimal rewards which allows the use of a reward function \emph{internal} to the agent that is potentially different from the \emph{objective} (or task-specifying) reward function. They showed that good choices of internal reward functions can mitigate agent limitations.\footnote{Others have developed methods for the design of rewards under alternate settings. Potential-based Reward Shaping~\cite{ng1999policy,asmuth2008potential} offers a space of reward functions with the property that an optimal policy for the original reward function remains an optimal policy for each reward function in the space. This allows the designer to heuristically pick a reward function for other properties, such as impact on learning speed. In some cases, reward functions are simply unknown, in which case inverse-RL~\cite{ng2000algorithms} has been used to infer reward functions from expert behavior. Yet others have explored the use of queries to a human expert to elicit preferences~\cite{chajewska2000making}.} \citet{sorg2010internal} proposed PGRD as a specific algorithm that exploits the insight that {for planning agents a reward function implicitly specifies a policy}, and adapts policy gradient methods to search for good rewards for UCT-based agents. As discussed in the Introduction, PGRD has limitations that we address next through the introduction of PGRD-DL. 

\section{PGRD-DL: Policy-Gradient for Reward Design with Deep Learning}

Understanding PGRD-DL requires understanding three major components. First, PGRD-DL uses UCT differently than usual in that it employs an internal reward function that is the sum of a reward-bonus and the usual objective reward (in ATARI games, the change in score). Second, there is a CNN-based parameterization of the reward-bonus function. Finally, there is a gradient procedure to train the CNN-parameters. We describe each in turn. 

\vspace*{0.5em}
{\bf \noindent UCT with internal rewards.}
As described in the Related Work section above, in extending a trajectory during planning, 
UCT computes a score that combines a UCB-based \emph{exploration term} that encourages sampling infrequently sampled actions with an \emph{exploitation term} computed from the trajectories sampled thus far. A full $H$-length trajectory is a sequence of state-action pairs: $s_{0}a_{0}s_{1}a_{1}...s_{H-1}a_{H-1}$. UCT estimates the exploitation-term value of a state, action, depth tuple $(s,a,d)$ as the average return obtained after experiencing the tuple (non-recursive form):
\begin{align}
Q(s,a,d)=\sum_{i=1}^{N} \frac{I_{i}(s,a,d)}{n(s,a,d)} \sum_{h=d}^{H-1}  \gamma^{h-d}R(s_{h}^{i}, a_{h}^{i}) \label{estimated_q}
\end{align}
where $N$ is the number of trajectories sampled, $\gamma$ is the discount factor, $n(s,a,d)$ is the number of times tuple $(s,a,d)$ has been sampled, $I_{i}(s,a,d)$ is 1 if $(s,a,d)$ is in the $i^{th}$ trajectory and 0 otherwise, $s_{h}^{i}$ is the $h^{th}$ state in the $i^{th}$ trajectory and $a_{h}^{i}$ is the $h^{th}$ action in the $i^{th}$ trajectory. 

The difference between the standard use of UCT and its use in PGRD-DL is in the choice of the reward function in Equation~\ref{estimated_q}. To emphasize this difference we use the notation $R^O$ to denote the usual \textit{objective} reward function, and $R^I$ to denote the new \textit{internal} reward function. 
Specifically, we let 
\begin{align}
R^{I}(s,a;\theta)=\mathrm{CNN}(s,a;\theta) + R^{O}(s,a) \label{basic_r}
\end{align}
where the reward-bonus that is added to the objective reward in computing the internal reward is represented via a multi-layered convolution neural network, or CNN, mapping from state-action pairs to a scalar; $\mathbf{\theta}$ denotes the CNN's parameters, and thus the reward-bonus parameters.  
To denote the use of internal rewards in Equation~\ref{estimated_q} and to emphasize its dependence on the parameters $\theta$, we will hereafter denote the Q-value function as $Q^I(\cdot,\cdot,\cdot;\theta)$. Note that the reward bonus in Equation~\ref{basic_r} is distinct from (and does not replace) the exploration bonus used by UCT during planning.

\vspace*{0.5em}
{\bf \noindent CNN parameterization of reward-bonuses.} PGRD-DL is capable of using any kind of feed-forward neural network to represent the reward bonus functions. The Experiment Setup section below defines the specific convolution network used in this work. %The specifics of this are defined in the Experiment Setup section below.
%(see Figure~\ref{fig:network} for a visualization).

\vspace*{0.5em}
{\bf \noindent Gradient procedure to update reward-bonus parameters.}
When UCT finishes generating the specified number of trajectories (when planning is complete), the greedy action is
\begin{align}
a=\arg\max_{b} Q^I(s,b,0;\theta)
\end{align}
where the action values of the current state are $Q^I(s,.,0;\theta)$. To allow for gradient calculations during training of the reward-bonus parameters, the
UCT agent executes actions according to a softmax distribution given the estimated action values (the temperature parameter is omitted):
\begin{align}
\mu(a|s; \theta)=\frac{\exp Q^I(s,a,0;\theta)}{\sum_{b}\exp Q^I(s,b,0;\theta)}, \label{policy}
\end{align}
where $\mu$ denotes the UCT agent's policy.

Even though internal rewards are used to determine the UCT policy, only the task-specifying objective reward $R^{O}$ is used to determine how well the internal reward function is doing. In other words, 
the performance of UCT with the internal reward function is measured over the experience sequence that consists of the actual executed actions and visited states: $h_{T}$= $s_{0}a_{0}s_{1}a_{1}...s_{T-1}a_{T-1}$ in terms of objective-return $u(.)$:
\begin{align}
u(h_{T})=\sum_{t=0}^{T-1}R^{O}(s_{t},a_{t}) \label{return}
\end{align}
where $s_{t}$ and $a_{t}$ denote the actual state and action at time $t$.
%, and $R^{O}(.)$ is the objective reward.
Here we assume that all tasks are episodic, and the maximum length of an experience sequence is $T$.
The expected objective-return is a function of the reward bonus function parameters $\theta$ because the policy $\mu$ depends on $\theta$, and in turn the policy determines the distribution over state-action sequences experienced. 

The PGRD-DL objective in optimizing reward bonus parameters is maximizing the expected objective-return of UCT:
\begin{align}
\theta^{*}=\arg\max_{\theta} U(\theta)=\arg\max_{\theta} \mathop{\mathbb{E}}\{u(h_{T})|\theta\}. \label{utility}
\end{align}

The central insight of PGRD was to consider the reward-bonus parameters as policy parameters and apply stochastic gradient ascent to maximize the expected return. Previous applications of PGRD used linear functions of hand-coded state-action-features as reward-bonus functions. In this paper, we first applied PGRD with a multi-layer convolutional neural network to automatically learn features from raw perception as well as to adapt the non-linear reward-bonus parameters.  However, empirical results showed that the original PGRD could cause the CNN parameters to diverge and cause degenerate performance due to large variance in the policy gradient estimation. We therefore adapted a variance-reduction policy gradient method GARB (GPOMDP with Average Reward Baseline;~\citet{weaver2001optimal}) to solve this drawback of the original PGRD. GARB  optimizes the reward-bonus parameters to maximize the expected objective-return as follows:
\begin{align}
\nabla_{\theta}U(\theta)&=\nabla_{\theta}\mathop{\mathbb{E}}\{u(h_{T})|\theta\}\\
&= \mathop{\mathbb{E}}\Big\{ u(h_{T}) \sum_{t=0}^{T-1} \frac{\nabla_{\theta} \mu(a_{t} | s_{t};\theta)}{\mu(a_{t} | s_{t};\theta)} \Big\} \label{reinforce_1}
\end{align}
Thus, $u(h_{T}) \sum_{t=0}^{T-1} \frac{\nabla_{\theta} \mu(a_{t} | s_{t};\theta)}{\mu(a_{t} | s_{t};\theta)}$ is an unbiased estimator of the objective-return gradient. GARB computes an eligibility trace vector $\mathbf{e}$ to keep track of the gradient vector $\mathbf{g}$ at time $t$:   
\begin{align}
\mathbf{e_{t+1}} &= \beta \mathbf{e_{t}} +  \frac{\nabla_{\theta} \mu(a_{t} | s_{t};\theta)}{\mu(a_{t} | s_{t};\theta)}\\
\mathbf{g}_{t+1} &= \mathbf{g}_{t} + (r_t - b) \mathbf{e_{t+1}}
\end{align}
where $\beta \in [ 0,1)$ is a parameter controlling the bias-variance trade-off of the gradient estimation, $r_{t}=R^{O}(s_{t}, a_{t})$ is the immediate objective-reward at time $t$, $b$ is a reward baseline and it equals the running average of $r_{t}$, and $\mathbf{g_{T}}$ is the gradient estimate vector. We use the gradient vector to update the reward parameters $\theta$ when a terminal state is reached; at the end of the $j^{th}$ episode, $\mathbf{g_{T}}$ is used to update $\theta$ using an existing stochastic gradient based method, ADAM~\cite{kingma2014adam}, as described below.
% such as RMSProp or ADAM.

Since the reward-bonus function is represented as a feed-forward network, back-propagation can compute the gradient of the reward-bonus function parameters, i.e. $\frac{\nabla_{\theta} \mu(a_{t} | s_{t};\theta)}{\mu(a_{t} | s_{t};\theta)}$, efficiently. In order to apply BackProp, we need to compute the derivative of policy $\mu$ with respect to the reward $R^{I}(s_{h}^{i}, a_{h}^{i}; \theta)$ in the $i^{th}$ sampling trajectory at depth $h$ in UCT planning:
\begin{align}
\delta^{r(i,h)}_{t}   &= \frac{1}{\mu(a_{t}|s_{t})} \frac{\partial \mu(a_{t}|s_{t}) }{\partial R^{I}(s_{h}^{i}, a_{h}^{i})}\\
&= \frac{1}{\mu(a_{t}|s_{t})}  \sum_{b} \frac{\partial \mu(a_{t}|s_{t}) }{\partial Q^I(s_{t}, b, 0)} \frac{\partial Q^I(s_{t}, b, 0)}{\partial R^{I}(s^{i}_{h}, a^{i}_{h})}\\
 &= \sum_{b} (I(a_{t} = b) - \mu(b|s_{t}))  \frac{I_{i}(s_{t},b,0)}{n(s_{t},b,0)} \gamma^{h}
\end{align}
where $I(a_{t} = b) = 1$ if $a_{t}$ equals $b$ and 0 otherwise. Thus the derivative of any parameter $\theta_{k}$ in the reward parameters can be represented as:
\begin{align}
\delta^{\theta_{k}}_{t}   &= \sum_{i,h} \delta^{r(i,h)}_{t} \frac{\partial R^{I}(s^{i}_{h}, a^{i}_{h})}{\partial \theta_{k}} 
\end{align}
where $\frac{\partial R^{I}(s^{i}_{h}, a^{i}_{h})}{\partial \theta_{k}} $ is determined by $(s^{i}_{h}, a^{i}_{h})$ and the CNN, and can be computed efficiently using standard BackProp.

\vspace*{0.5em}
{\bf \noindent What does the PGRD-DL learn?} We emphasize that the only thing that is learned from experience during repeated application of the PGRD-DL planning procedure is the reward-bonus function. All the other aspects of the PGRD-DL procedure remain fixed throughout learning. 
%The reward-bonus function is thus the locus of learning that 

\section{Experimental Setup: Evaluating PGRD-DL on ATARI Games}

We evaluated PGRD-DL on 25 ATARI games (Table~\ref{table:mainresults}). 
All the ATARI games involve controlling a game agent in a 2-D space, but otherwise have very different dynamics.

\vspace*{0.5em}
{\bf \noindent UCT objective reward and planning parameters.}
As is standard in RL work on ATARI games, we take the objective reward for each state to be the difference in the game score between that state and the previous state. We rescale the objective reward: assigning +1 for positive rewards, and -1 for negative rewards. A game-over state or life-losing state is considered a terminal state in UCT planning and PGRD training. Evaluation trajectories only consider game-over states as terminal states.

All UCT baseline agents in our experiments sample 100 trajectories of depth 100 frames\footnote{Normally UCT does planning for every visited state. However, for some states in ATARI games the next state and reward is the same no matter which action is chosen (for example, when the Q*Bert agent is falling) and so UCT planning is a waste of computation. In such states our agents do not plan but instead choose a random action.}.
The UCB-parameter is 0.1 and the discount factor $\gamma=0.99$.  
Following \citet{mnih2015human} we use a simple frame-skipping technique to save computations: the agent selects actions on every $4^{th}$ frame instead of every frame, and the last action is repeated on skipped frames. We did not apply PGRD to ATARI games in which UCT already achieves the highest possible score, such as Pong and Boxing.
%We do not apply PGRD to ATARI games if UCT achieves the highest scores, such as Pong and Boxing.

\vspace*{0.5em}
{\bf \noindent Screen image preprocessing.}
The last four game screen images are used as input for the CNN. The 4 frames are stacked in channels.
The game screen images (210 $\times$ 160) are downsampled to 84 $\times$ 84 pixels and gray-scaled. Each image is further preprocessed by pixel-wise mean removal. The pixel-wise mean is calculated over ten game trajectories of a uniformly-random policy.

%\begin{figure}[tb]
%  \centering
%      \includegraphics[width=0.4\textwidth]{figures/network}
%      
%  \caption{The network architecture for learning and representing the reward-bonus function. The network %consists of 3 hidden layers: 2 convolution layers and 1 fully-connected layer. See the text for details. %\label{fig:network}}
%\end{figure}

\vspace*{0.5em}
{\bf \noindent Convolutional network architecture.}
The same network architecture is used for all games.
%and is shown in Figure~\ref{fig:network}.
The network consists of 3 hidden layers. The first layer convolves 16, 8$\times$8 filters with stride 4. The second hidden layer convolves 32, 4$\times$4 filters with stride 2. The third hidden layer is a full-connected layer with 256 units. Each hidden layer is followed by a rectifier nonlinearity. The output layer has one unit per action.
% TODO: reference for ADAM (Done): kingma2014adam

\vspace*{0.5em}
{\bf \noindent PGRD-DL learning parameters.}
After computing the gradients of CNN parameters, we use ADAM to optimize the parameters of the CNN. ADAM is an adaptive stochastic gradient optimization method to train deep neural networks \cite{kingma2014adam}. We use the default hyper-parameters of ADAM\footnote{A discount factor of 0.9 to compute the accumulated discount sum of the first moment vector and 0.999 for the second order moment vector.}. We set $\beta=0.99$ in GARB. Thus the only remaining hyper-parameter is the learning rate for ADAM, which we selected from a candidate set \{$10^{-3}$, $10^{-4}$, $10^{-5}$, $10^{-6}$\} by identifying the rate that produced the greatest sum of performance improvements after training 1000 games on Ms.\ Pacman and Q*Bert.  The selected learning rate of $10^{-4}$ served as the initial learning rate for PGRD-DL. The learning rate was then lowered during learning by dividing it by 2 every 1000 games.

\section{Experimental Results }

We allowed PGRD-DL to adapt the bonus reward function for at most 5000 games or at most 1 million steps, whichever condition was satisfied first. For each game, the final learned reward bonus function was then evaluated in UCT play (using the same depth and trajectory parameters specified above) on 20 ``evaluation'' games, during which the reward function parameters were held fixed and UCT selected actions greedily according to the action value estimate of root nodes. Note that even though UCT chose actions according to action values greedily, there was still stochasticity in UCT's behavior because of the randomized tree expansion procedure of UCT.  

\begin{table*}[htb]
\caption{Performance comparison of different UCT planners. The {\em $R^{O}$} columns denote UCT agents planning with objective rewards: {\em $R^{O}$} is depth 100 with 100 trajectories, {\em $R^{O}$(deeper)} is depth 200 with 100 trajectories, and {\em $R^{O}$(wider)} is depth 100 with 200 trajectories. The {\em $R^{I}$} column shows UCT's performance with internal rewards learned by PGRD-DL, with depth 100 and 100 trajectories.  The table entries are mean scores over 20 evaluation games, and in parentheses are the standard errors of these means.  The last two columns show the performance ratio of $R^{I}$ compared to $R^{O}$ and {\em max($R^{O}$(deeper),$R^{O}$(wider))} (denoted max($R^{O}$(deeper,wider)) in the last column of the table). The $R^{I}$, $R^{O}$(deeper), and $R^{O}$(wider) agents take approximately equal time per decision.}
\label{table:mainresults}
{\small
\begin{center}
    \begin{tabular}{lcccccc}
    \toprule
    & \multicolumn{4}{c}{\em Mean Game Score (standard error)}
    & \multicolumn{2}{c}{\em Mean Game Score Ratios}\\
\cmidrule(lr){2-5} \cmidrule(lr){6-7}
\em    Game & \em $R^{O}$ & \em $R^{I}$ & \em $R^{O}$(deeper) & \em $R^{O}$ (wider) & \em $R^{I}$ / $R^{O}$ & \em $R^{I}$ / max($R^{O}$(deeper,wider)) \\
    \midrule 
Alien & 2246 (139) & \maxscore{12614} (1477) & 2906 (387) & 1795 (218) & 5.62 & 4.34 \\ 
Amidar & 152 (13) & \maxscore{1122} (139) & 204 (20) & 144 (16) & 7.39 & 5.50 \\ 
Assault & 1477 (36) & 1490 (32) & 1495 (42) & \maxscore{1550} (59) & 1.01 & 0.96 \\ 
Asterix & 11700 (3938) & 60353 (19902) & \maxscore{99728} (16) & 77211 (10377) & 5.16 & 0.61 \\ 
BankHeist & 226 (13) & 248 (13) & 262 (17) & \maxscore{284} (19) & 1.10 & 0.88 \\ 
BattleZone & 8550 (879) & \maxscore{17450} (1501) & 13800 (1419) & 8450 (1274) & 2.04 & 1.26 \\ 
BeamRider & 2907 (322) & 2794 (232) & \maxscore{2940} (537) & 2526 (333) & 0.96 & 0.95 \\ 
Berzerk & 467 (25) & 460 (26) & \maxscore{506} (48) & 458 (35) & 0.99 & 0.91 \\ 
Breakout & 48 (14) & \maxscore{746} (24) & 516 (38) & 79 (30) & 15.47 & 1.45 \\ 
Carnival & 3824 (240) & \maxscore{5610} (678) & 3827 (173) & 3553 (218) & 1.47 & 1.47 \\ 
Centipede & \maxscore{4450} (236) & 3987 (185) & 2771 (231) & 4076 (325) & 0.90 & 0.98 \\ 
DemonAttack & 5696 (3316) & \maxscore{121472} (201) & 72968 (13590) & 67166 (11604) & 21.32 & 1.66 \\ 
MsPacman & 4928 (513) & \maxscore{10312} (781) & 6259 (927) & 4967 (606) & 2.09 & 1.65 \\ 
Phoenix & 5833 (205) & \maxscore{6972} (371) & 5931 (370) & 6052 (330) & 1.20 & 1.15 \\ 
Pooyan & 11110 (856) & \maxscore{20164} (1015) & 13583 (1327) & 13106 (1605) & 1.81 & 1.48 \\ 
Q*Bert & 2706 (409) & \maxscore{47599} (2407) & 6444 (1020) & 4456 (688) & 17.59 & 7.39 \\ 
RiverRaid & 3406 (149) & \maxscore{5238} (335) & 4165 (306) & 4254 (308) & 1.54 & 1.23 \\ 
RoadRunner & 8520 (3330) & \maxscore{32795} (4405) & 12950 (3619) & 7217 (2758) & 3.85 & 2.53 \\ 
Robotank & 2 (0.26) & 3 (0.38) & \maxscore{6} (0.84) & 1 (0.33) & 1.66 & 0.47 \\ 
Seaquest & 422 (19) & \maxscore{2023} (251) & 608 (41) & 518 (45) & 4.79 & 3.33 \\ 
SpaceInvaders & 1488 (114) & 1824 (88) & \maxscore{2154} (142) & 1516 (166) & 1.23 & 0.85 \\ 
StarGunner & 21050 (1507) & \maxscore{826785} (3865) & 33000 (4428) & 22755 (1294) & 39.28 & 25.05 \\ 
UpNDown & 127515 (10628) & 103351 (5802) & 109083 (9949) & \maxscore{144410} (38760) & 0.81 & 0.72 \\ 
VideoPinball & 702639 (17190) & 736454 (23411) & \maxscore{845280} (88556) & 779624 (90868) & 1.05 & 0.87 \\ 
WizardOfWor & 140475 (7058) & \maxscore{198495} (225) & 152886 (7439) & 149957 (7153) & 1.41 & 1.30 \\ 
    \bottomrule
    \end{tabular}
\end{center}
}
\end{table*}

\subsection{Learned reward bonuses improve UCT}
Table~\ref{table:mainresults} shows the performance of UCT using the objective game-score-based reward ({\em $R^{O}$} column) and UCT with the learned reward bonus ({\em $R^{I}$} column). The values show the mean scores over 20 evaluation games and the numbers in parentheses are the standard errors of these means.  

The results show that  PGRD-DL learns reward bonus functions that improve  UCT significantly on 18 out of 25 games.
(all except Assault, BankHeist, BeamRider, Berzerk, Centipede, UpNDown and VideoPinball).
The mean scores of UCT with learned rewards are significantly higher than UCT using the game-score-based reward for these 18 games. The $R^{I}/R^{O}$ column displays the ratio of the mean scores in column $R^{I}$ to those in column $R^{O}$, and so a ratio over 1 implies improvement. These ratio values are plotted as a blue (dashed) curve in Figure~\ref{fig:results}; the learned rewards improve UCT by a ratio of more than $10$ on $4$ games (StarGunner, DemonAttack, Q*Bert and Breakout), and a ratio between 2 and $10$ on $7$ games (Amidar, Alien, Asterix, Seaquest, RoadRunner, MsPacman and BattleZone). 
% Figure~\ref{fig:results} shows the sorted performance ratio between  {\em UCT+$R^{O}$} and {\em UCT+$R^{I}$}. 

\subsection{Comparison considering computational cost}
The previous results establish that planning using the learned internal reward bonus can improve the performance of UCT on many ATARI games. However there is some computational overhead incurred in using the deep network to compute the internal reward bonuses (300 ms and 200 ms for UCT with $R^{I}$ and $R^{O}$ respectively). Is it worth spending the additional computational resources to compute the internal reward, as opposed to simply planning more deeply or broadly with the objective reward? 
%The above results ignore the computational overhead incurred by using the reward bonuses. 
We consider now the performance of two additional baselines that use additional computation to improve UCT without reward bonuses. 
%consider two cases to use such computational overhead to improve UCT without reward bonuses. 
The first baseline, {\em deeper UCT}, plans to a depth of 200 frames instead of 100 frames (the number of trajectories is still 100 trajectories). The second baseline, {\em wider UCT},  samples 200 trajectories to depth 100 (the depth is still 100 frames). Both deeper UCT and wider UCT use about the same computational overhead as the UCT agent with reward bonuses that samples 100 trajectories to depth 100; the time per decision of deeper or wider UCT is slightly greater than UCT with reward bonuses. 
%(300 ms).

\begin{figure}[htb]
  \centering
      \includegraphics[width=0.95\columnwidth]{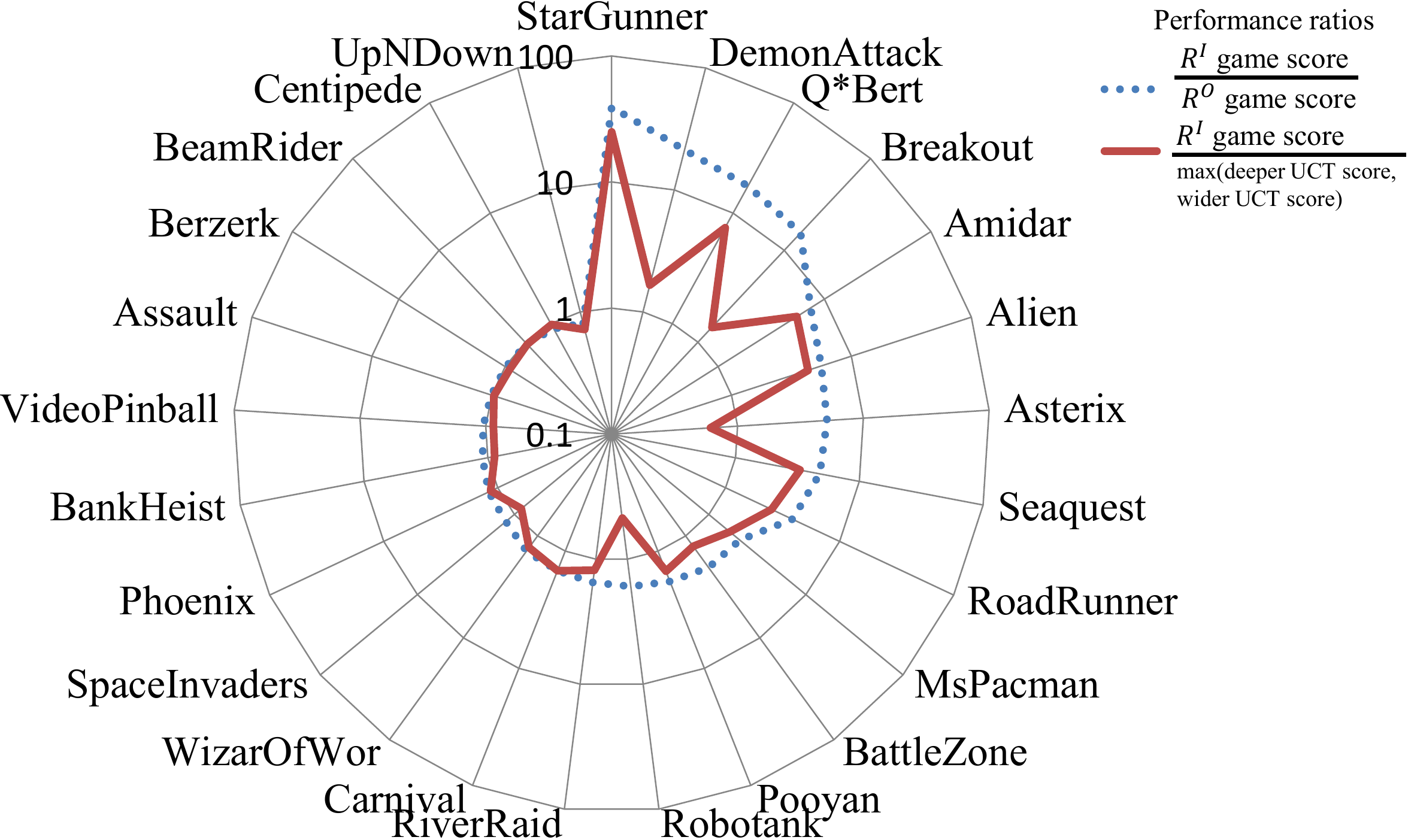}
      
  \caption{Performance comparison summary. The blue (dashed) curve shows the ratio of the mean game score obtained by UCT planning with the PGRD-DL-adapted internal reward $R^{I}$, and the mean game score obtained by UCT planning with the objective reward $R^{O}$. The red (solid) curve shows the ratio of the mean game score obtained by UCT with $R^{I}$, and the mean game score of UCT planning with $R^{O}$ with either deeper or more (wider) trajectories (whichever yields the higher score). 
  UCT with the internal reward bonus outperforms the baseline if the ratio value lies outside the circle with radius 1. Games are sorted according to {\em $R^{I}/R^{O}$}. \label{fig:results}}
\end{figure}

The mean scores of the deeper UCT and wider UCT are summarized in Table~\ref{table:mainresults}. %We take the higher of these two scores to compare against UCT using the learned internal reward $R^{I}$. 
We take the higher of the mean scores of the deeper and wider UCTs as a useful assessment of performance obtainable using the computational overhead of reward bonuses for better planning, and compare it to the performance of UCT using the learned internal reward $R^{I}$.
The last column in Table~\ref{table:mainresults} displays the ratio of the mean scores in column $R^{I}$ to the higher of the wider UCT and deeper UCT scores, and this ratio appears as the red line in Figure~\ref{fig:results}.
Among the 18 games in which reward bonuses improve UCT, reward bonuses outperform even the better of deeper or wider UCT agents in 15 games. These results show that the additional computational resources required to compute the reward bonuses may be better spent in this way than using those resources for more extensive planning. 

%can work better than using the computational overhead in planning for those 15 games. 
%The performance ratio values are plotted using red curve in Figure~\ref{fig:results}. 

\subsection{The nature of the learned reward-bonuses}

What kinds of state and action discriminations does the reward bonus function learn? We consider now a simple summary visualization of the reward bonuses over time, as well as specific examples from the games Ms.\ Pacman and Q*Bert.  
The key conclusion from these analyses is that PGRD-DL learns useful game-specific features of state that help UCT planning, in part by mitigating the challenge of delayed reward.

\begin{figure}[tbh]
\centering
 \includegraphics[width=0.95\columnwidth]{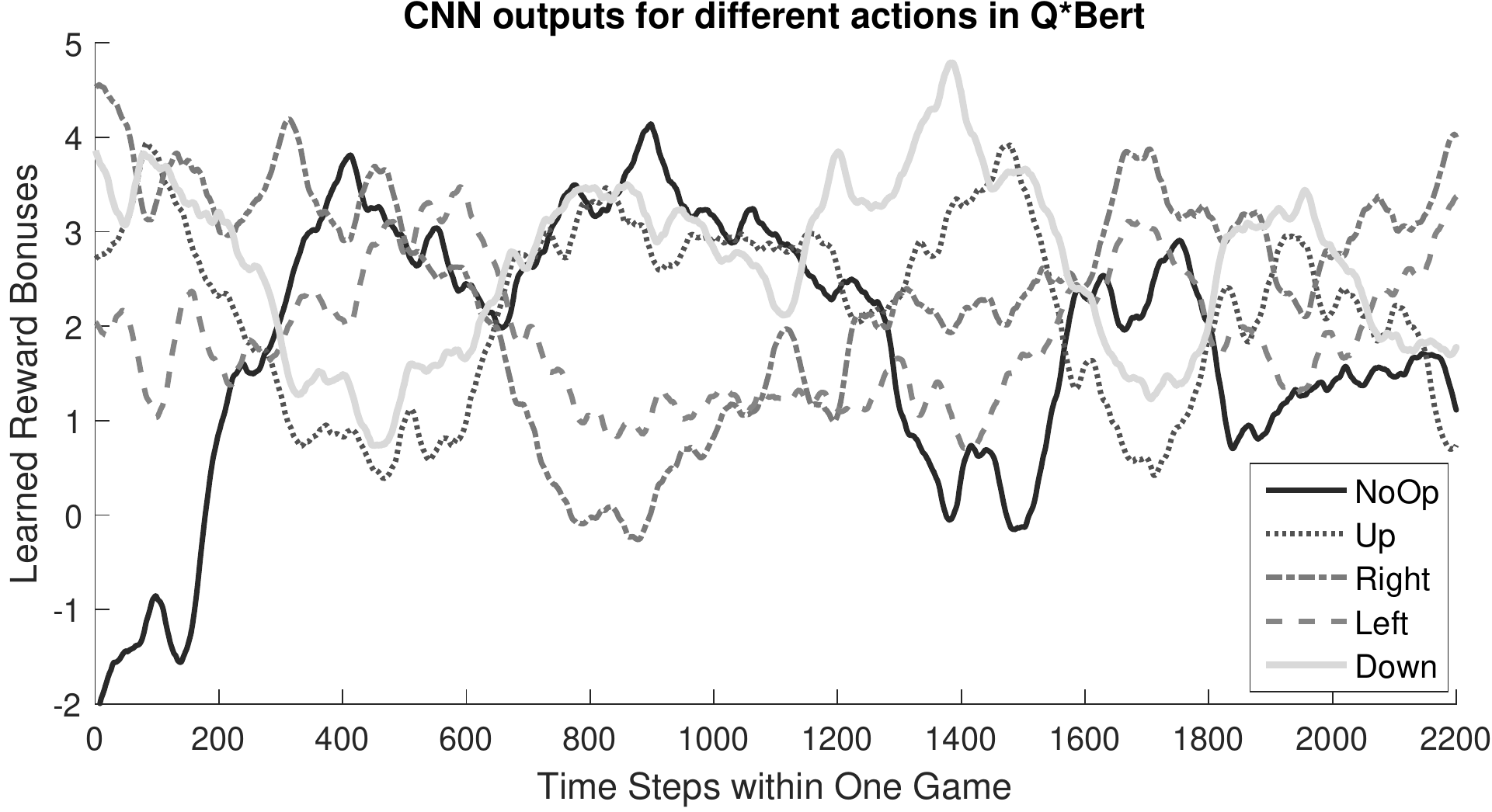}
\caption{The learned reward bonuses for each of the five actions in Q*Bert for the states 
experienced during one game play. It is visually clear that different actions have the largest reward-bonus in different states.}
\label{fig:cnn-output}
\end{figure}

\vspace*{0.5em}
{\bf \noindent Visualizing the dynamically changing reward bonus across states experienced in game play.} Consider first how the learned reward bonus for each action changes as a function of state.  Figure~\ref{fig:cnn-output} shows the varying learned reward bonus values for each of the five actions in Q*Bert for the states experienced during one game play. The action with the highest (and lowest) reward bonus changes many times over the course of the game.
The relatively fine-grained temporal dynamics of the reward bonuses throughout the game, and especially the change in relative ordering of the actions, provides  support for the claim that the learned reward makes game-specific state discriminations---it is not simply unconditionally increasing or decreasing rewards for particular actions, which would have resulted in mostly flat lines across time in Figure~\ref{fig:cnn-output}. We now consider specific examples of the state discriminations learned. 

\begin{figure}
\centering
\includegraphics[width=0.45\columnwidth]{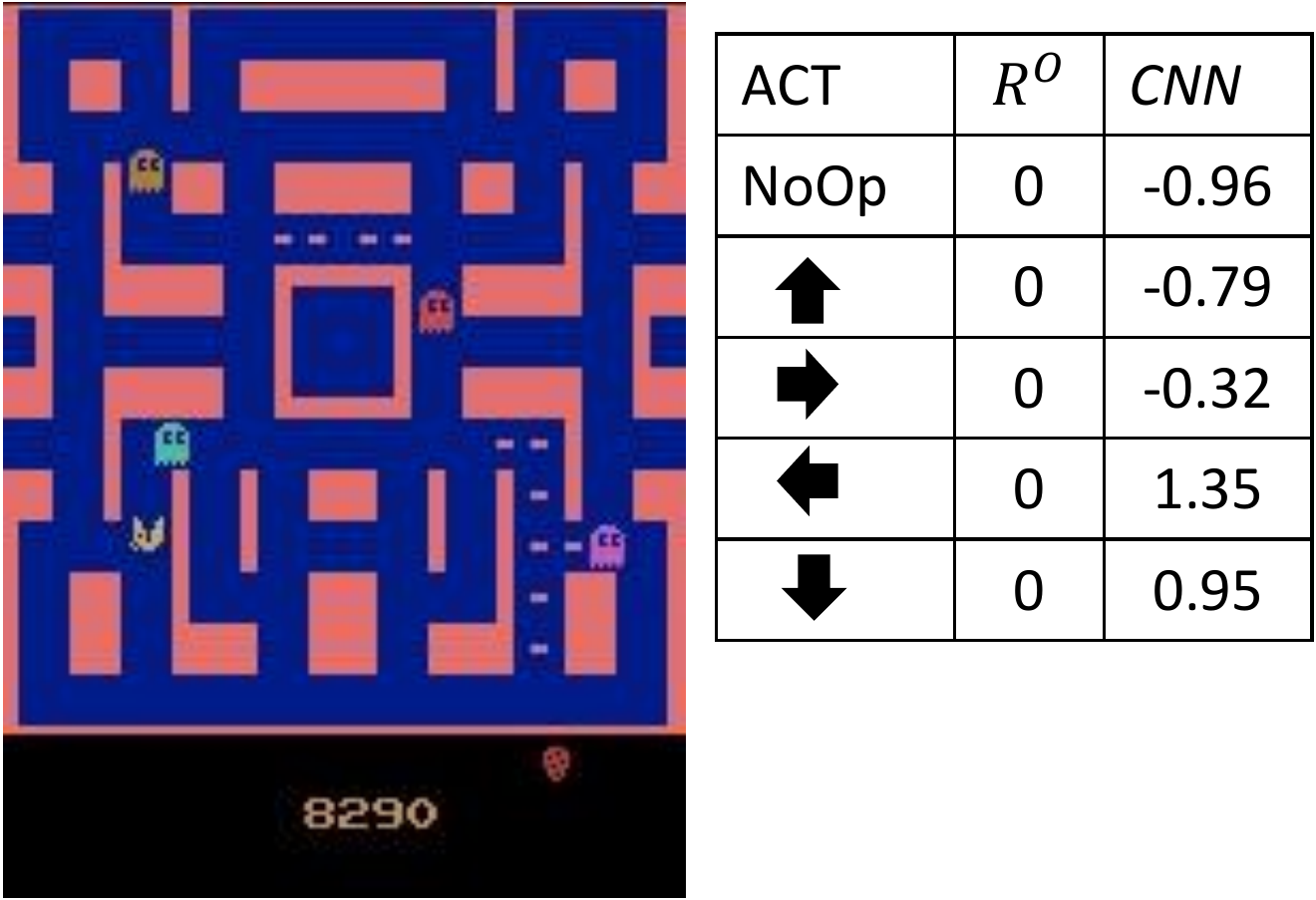}~
\includegraphics[width=0.45\columnwidth]{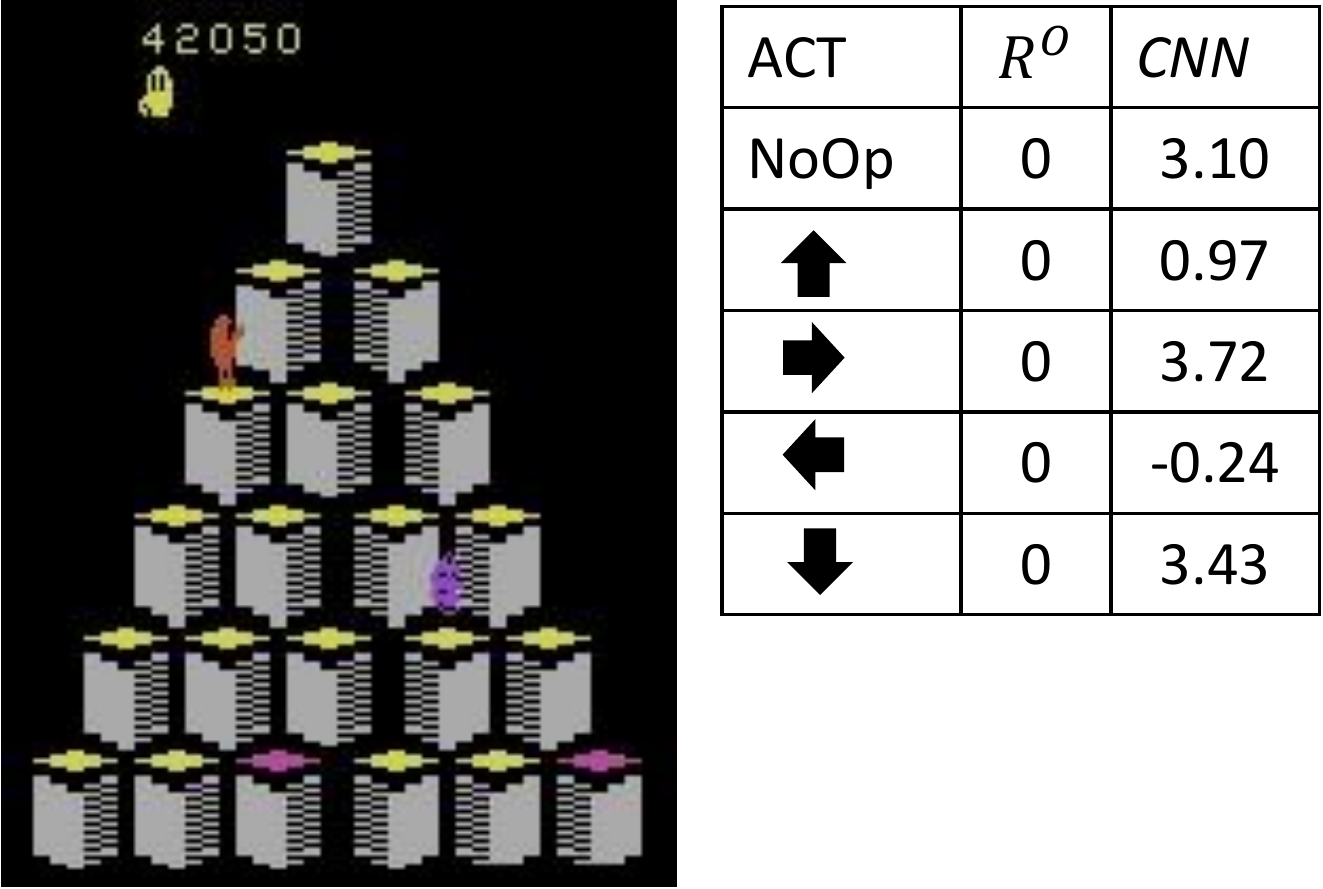}
\caption{\label{fig:vis}
Examples of how the reward bonus function represented by the CNN learns to encourage actions that avoid delayed bad outcomes. {\em Left}: A state in Ms.\ Pacman where the agent will encounter an enemy if it continues moving up. The learned reward bonus (under ``CNN'' in the small table) gives positive reward for actions taking the agent away and negative reward for actions that maintain course; the objective game score does not change (and so $R^O$ is zero). {\em Right}: A state in Q*Bert where the agent could fall off the pyramid if it moves left and so left is given a negative bonus and other actions are given positive bonuses. The objective reward $R^O$ is zero and indeed continues to be zero as the agent falls.}
\end{figure}

\vspace*{0.5em}
{\bf \noindent Examples of learned reward bonuses that capture delayed reward consequences.} 
In the game Ms.\ Pacman, there are many states in which it is important to choose a movement direction that avoids subsequent encounters with enemies (and loss of a ``Pacman life'').  These choices may not yield differences in immediate reward and are thus  examples of delayed reward consequences. Similarly, in Q*Bert, falling from the pyramid is a bad outcome but the falling takes many times steps before the ``life'' is lost and the episode ends.   
These consequences could in principle be taken into account by UCT planning with sufficient trajectories and depth. But we have discovered through observing the game play and examining specific bonus rewards that PGRD-DL learns reward bonuses that encourage action choices avoiding future enemy contact in Ms.\ Pacman and falling in Q*Bert (see Figure~\ref{fig:vis}; the figure caption provides detailed descriptions.) The key lesson here is that PGRD-DL is learning useful and interesting game-specific state discriminations for a reward bonus function that mitigates the problem of delayed objective reward.

\section{Conclusions}
In this paper we introduced a novel approach to combining Deep Learning and Reinforcement Learning by using the former to learn good reward-bonus functions from experience to improve the performance of UCT on ATARI games. Relative to the state-of-the art in the use of PGRD for reward design, we also provided the first example of automatically learning features of raw perception for use in the reward-bonus function, the first use of nonlinearly parameterized reward-bonus functions with PGRD, and provided empirical results on the most challenging domain of application of PGRD thus far. Our adaptation of PGRD uses a variance-reducing gradient procedure to stabilize the gradient calculations in the multi-layer CNN. Our empirical results showed that PGRD-DL learns reward-bonus functions that can significantly improve the performance of UCT, and furthermore that the learned reward-bonus functions can mitigate the computational limitations of UCT in interesting ways. While our empirical results were limited to ATARI games, PGRD-DL is fairly general and we expect it to generalize it to other types of domains. Combining more sophisticated DL architectures, e.g., LSTM~\cite{hochreiter1997long}, with RL in learning reward-bonus functions remains future work.

%\begin{small}
\noindent
{\bf Acknowledgments.} This work was supported by NSF grant IIS-1526059. Any opinions, findings, conclusions, or recommendations expressed here are those of the authors and do not necessarily reflect the views of the sponsor.
%\end{small}

\appendix
\begin{small}
\bibliographystyle{named}
\bibliography{reference}
\end{small}
\end{document}